\documentclass[10pt,twocolumn,letterpaper]{article}

\usepackage{wacv}
\usepackage{times}
\usepackage{epsfig}
\usepackage{graphicx}
\usepackage{amsmath}
\usepackage{amssymb}

\usepackage{amsfonts, libertine, verbatim, subcaption, multirow, float}
\usepackage{threeparttable}
\usepackage[pagebackref=true,breaklinks=true,letterpaper=true,colorlinks,bookmarks=false]{hyperref}

\newcommand{\ii}{{\bf i}}
\newcommand{\cc}{{\bf c}}
\newcommand{\oo}{{\bf o}}
\newcommand{\ff}{{\bf f}}
\newcommand{\hh}{{\bf h}}
\newcommand{\xx}{{\bf x}}

\newcommand{\bb}{{\bf b}}

\graphicspath{{Images/}}

\wacvfinalcopy % *** Uncomment this line for the final submission

\def\wacvPaperID{60} % *** Enter the wacv Paper ID here

\ifwacvfinal\fi
\begin{document}
	%%%%%%%%% TITLE
	\title{CSVideoNet: A Real-time End-to-end Learning Framework for High-frame-rate Video Compressive Sensing}
	
	\author{Kai XU \hspace{2cm} Fengbo Ren \\
		School of Computing, Informatics, and Decision Systems Engineering \\
		Arizona State University, Tempe AZ 85281 \\
		{\tt\small \{kaixu, renfengbo\}@asu.edu}
	}
	
	\maketitle
	\ifwacvfinal\thispagestyle{empty}\fi
	
	\begin{abstract}
		This paper addresses the real-time encoding-decoding problem for high-frame-rate video compressive sensing (CS). Unlike prior works that perform reconstruction using iterative optimization-based approaches, we propose a non-iterative model, named ``CSVideoNet", which directly learns the inverse mapping of CS and reconstructs the original input in a single forward propagation. To overcome the limitations of existing CS cameras, we propose a multi-rate CNN and a synthesizing RNN to improve the trade-off between compression ratio (CR) and spatial-temporal resolution of the reconstructed videos. The experiment results demonstrate that CSVideoNet significantly outperforms state-of-the-art approaches. Without any pre/post-processing, we achieve a 25dB Peak signal-to-noise ratio (PSNR) recovery quality at 100x CR, with a frame rate of 125 fps on a Titan X GPU. Due to the feedforward and high-data-concurrency natures of CSVideoNet, it can take advantage of GPU acceleration to achieve three orders of magnitude speed-up over conventional iterative-based approaches. We share the source code at https://github.com/PSCLab-ASU/CSVideoNet.
	\end{abstract}
	
	\section{Introduction}
	High-frame-rate cameras are capable of capturing videos at frame rates over 100 frames per second (fps). These devices were originally developed for research purposes, e.g., to characterize events that occur at a rate that traditional cameras are incapable of recording in physical and biological science. Some high-frame-rate cameras, such as Photron SA1, SA3, are capable of recording high resolution still images of ephemeral events such as a supersonic flying bullet or an exploding balloon with negligible motion blur and image distortion artifacts. However, due to the complex sensor hardware designed for high sampling frequency, these types of equipment are extremely expensive (over tens of thousand dollars for one camera). The high cost limits the field of their applications. Furthermore, the high transmission bandwidth and the large storage space associated with the high frame rate challenges the manufacture of affordable consumer devices. For example, true high-definition-resolution (1080p) video cameras at a frame rate of 10k fps can generate about 500 GB data per second, which imposes significant challenges on existing transmission and storage techniques. Also, the high throughput raises energy efficiency a big concern. For example, ``GoPro 5" can capture videos at 120 fps with 1080p resolution. However, the short battery life (1-2 hours) has significantly narrowed their practical applications.
	
	Traditional video encoder, e.g., H.264/MPEG-4, is composed of motion estimation, frequency transform, quantization, and entropy coding modules. From both speed and cost perspectives, the complicated structure makes these video encoder unsuitable for high-frame-rate video cameras. Alternatively, compressive sensing (CS) is a much more hardware-friendly acquisition technique that allows video capture with a sub-Nyquist sampling rate. The advent of CS has led to the emergence of new image devices, e.g., single-pixel cameras \cite{Duarte:CS}. CS has also been applied in many practical applications, e.g., accelerating magnetic resonance imaging (MRI) \cite{Ma:MRI}. While traditional signal acquisition methods follow a sample-then-compress procedure, CS could perform compression along with sampling. The novel acquisition strategy has enabled low-cost on-sensor data compression, relieving the pain for high transmission bandwidth and large storage space. In the recent decade, many algorithms have been proposed \cite{Candes:robust, Needell:cosamp, Amir:ISTA, Daubechies:IRWLS, Tropp:omp, Thomas:IHT, rebirth:li} to solve the CS reconstruction problem. Generally, these reconstruction algorithms are based on either optimization or greedy approaches using signal sparsity as prior knowledge. As a result, they all suffer from high computational complexity, which requires seconds to minutes to recover an image depending on the resolution. Therefore, these sparsity-based methods cannot satisfy the real-time decoding need of high-frame-rate cameras, and they are not appropriate for the high-frame-rate video CS application.
	
\begin{figure} 
		\centering
		\includegraphics[width=0.45\textwidth]{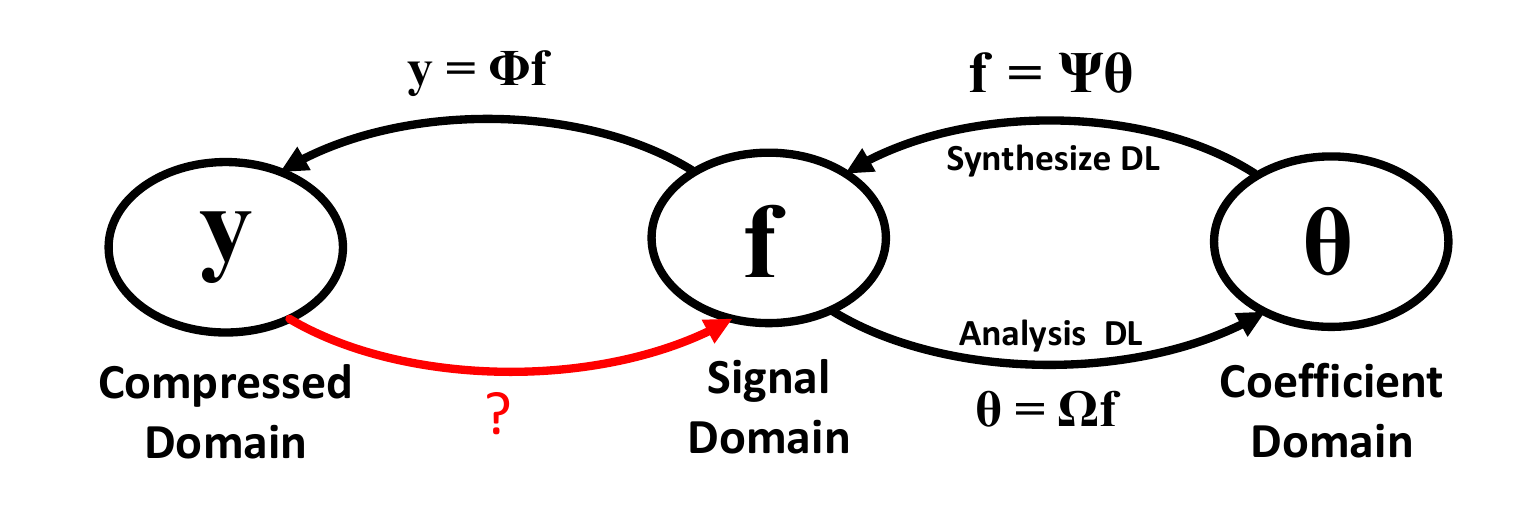}
		\caption{Illustration of domain transformations in CS. This work bridges the gap between compressed and signal domains.}
		\label{fig:CS}
		\vspace{-0.12in}
	\end{figure}

	The slow reconstruction speed of conventional CS approaches motivates us to directly model the inverse mapping from the compressed domain to original domain, which is shown in Figure~\ref{fig:CS}. Usually, this mapping is extremely complicated and difficult to model. However, the existence of massive unlabeled video data gives a chance to learn such a mapping using data-driven methods. In this paper, we design an enhanced Recurrent convolutional neural network (RCNN) to solve this problem.  RCNN has shown astonishingly good performance for video recognition and description \cite{Jeff:recurrent, Venugopalan:s2vt, Xu:show, srivastava:video}. However, conventional RCNNs are not well suited for video CS application, since they are mostly designed to extract discriminant features for classification related tasks. Simultaneously improving compression ratio (CR) and preserving visual details for high-fidelity reconstruction is a more challenging task. To solve this problem, we develop a special RCNN, called ``CSVideoNet", to extract spatial-temporal features, including background, object details, and motions, to significantly improve the compression ratio and recovery quality trade-off for video CS application over existing approaches.
	
	The contributions of this paper are summarized as follows:
	\begin{itemize}
		\item We propose an end-to-end and data-driven framework for video CS. The proposed network directly learns the inverse mapping from the compressed videos to the original input without additional pre/post-processing. To the best of our knowledge, there has been no published work that addresses this problem using similar methods.
		\item We propose a multi-level compression strategy to improve CR with the preservation of high-quality spatial resolution. Besides, we perform implicit motion estimation to improve temporal resolution. By combining both spatial and temporal features, we further improve the compression ratio and recovery quality trade-off without increasing much computational complexity.
		\item We demonstrate CSVideoNet outperforms the reference approaches not only in recovery quality but also in reconstruction speed because of its non-iterative nature. It enables real-time high-fidelity reconstruction for high-frame-rate videos at high CRs. We achieve state-of-the-art performance on the large-scale video dataset UCF-101. Specifically, CSVideoNet reconstructs videos at 125 fps on a Titan X GPU and achieves 25dB PSNR at a 100x CR.
	\end{itemize}
	
	\section{Related work}
	There have been many recovery algorithms proposed for CS reconstruction, which can  be categorized as follows:
	
	\textbf{Conventional Model-based CS Recovery}:
	In \cite{Baraniuk:LDS}, the authors model the evolution of scenes as a linear dynamical system (LDS). This model comprises two sub-models: the first is an observation model that models frames of video lying on a low-dimensional subspace; the second predicts the smoothly varied trajectory. The model performs well in stationary scenes, however, inadequate for non-stationary scenes. 
	
	In \cite{Yang:GMM}, the authors use Gaussian mixture model (GMM) to recover high-frame-rate videos, and the reconstruction can be efficiently computed as an analytical solution. The hallmark of the algorithm is that it adapts temporal compression rate based upon the complexity of the scene. The parameters in GMM are trained off-line and tuned during the recovery process.
	
	In \cite{Aswin:multiplexing}, the authors propose a multi-scale video recovery framework. It first obtains a low-resolution video preview with very low computational complexity, and then it exploits motion estimates to recover the full-resolution video by solving an optimization problem. In a similar work \cite{Fowler:motion}, the authors propose a motion-compensated and block-based CS reconstruction algorithm with smooth projected Landweber (MC-BCS-SPL). The motion vector is estimated from a reference and a reconstructed frame. The reconstructed video is derived from the combination of the low-resolution video and the estimated motion vector. The drawback of the two work is the requirement of specifying the resolution at which the preview frame is recovered, which requires prior knowledge of the object speed. Also, the recovery performance is highly dependent on the quality of motion estimation. To accurately estimate motion vector is a challenging task especially in high-frame-rate scenarios. The high computational cost further makes this model inadequate for reconstructing high-frame-rate videos.
	
	\textbf{Deep Neural Network (DNN) Based CS Recovery:}
	In \cite{Mousavi:DLRecovery}, the authors propose a stacked autoencoder to learn a representation of the training data and to recover test data from their sub-sampled measurements. Compared to the conventional iterative approaches, which usually need hundreds of iterations to converge, the feed-forward deep neural network runs much faster in the inference stage.
	
	In \cite{Kulkarni:ReconNet}, the authors propose a convolutional neural network, which takes CS measurements of an image as input and outputs an intermediate reconstruction. The intermediate output is fed into an off-the-shelf denoiser to obtain the final reconstructed image. The author shows the network is highly robust to sensor noise and can recover visually higher quality images than competitive algorithms at low CRs of 10 and 25. Both \cite{Mousavi:DLRecovery} and \cite{Kulkarni:ReconNet} are designed for image reconstruction, which only focus on spatial feature extraction. For video applications, temporal features between adjacent frames are also important. Therefore, the overlook of temporal correlation makes the image reconstruction algorithms inadequate for video applications. 
	
	In \cite{Iliadis:Video}, the authors propose a Video CS reconstruction algorithm based on a fully-connected neural network. This work focuses on temporal CS where multiplexing occurs across the time dimension. A 3D volume is reconstructed from 2D measurements by a feed-forward process. The author claims the reconstruction time for each frame can be reduced to about one second. The major drawback of this work is that the algorithm is based on a plain fully-connected neural network, which is not efficient in extracting temporal features.
	
	\section{Methodology}
	\subsection{Overview of the proposed framework for video CS}
	Two kinds of CS cameras are being used today. Spatial multiplexing cameras (SMC) take significantly fewer measurements than the number of pixels in the scene to be recovered. SMC has low spatial resolution and seeks to spatially super-resolve videos. In contrast, temporal multiplexing cameras (TMC) have a high spatial resolution but low frame-rate sensors. Due to the missing of inter frames, extra computation is needed for motion estimation. For these two sensing systems, either spatial or temporal resolution is sacrificed for achieving a better spatial-temporal trade-off. To solve this problem, we propose a new sensing and reconstruction framework, which combines the advantage of the two systems. The random video measurements are collected by SMC with very high temporal resolution. To compensate for the low spatial resolution problem in SMC, we propose a multi-CR strategy. The first \textit{key frame} in a group of pictures (GOP) is compressed with a low CR, and the remaining \textit{non-key frames} are compressed with a high CR. The spatial features in the key frame are reused for the recovery of the entire GOP due to the high inter-frame correlation in high-frame-rate videos. The spatial resolution is hence improved. The RNN extrapolates motion from high-resolution frames and uses it to improve the temporal resolution. Therefore, a better compression ratio and spatial-temporal resolution trade-off are obtained by the proposed framework.
	
\begin{figure*} 
		\centering
		\includegraphics[width=\textwidth]{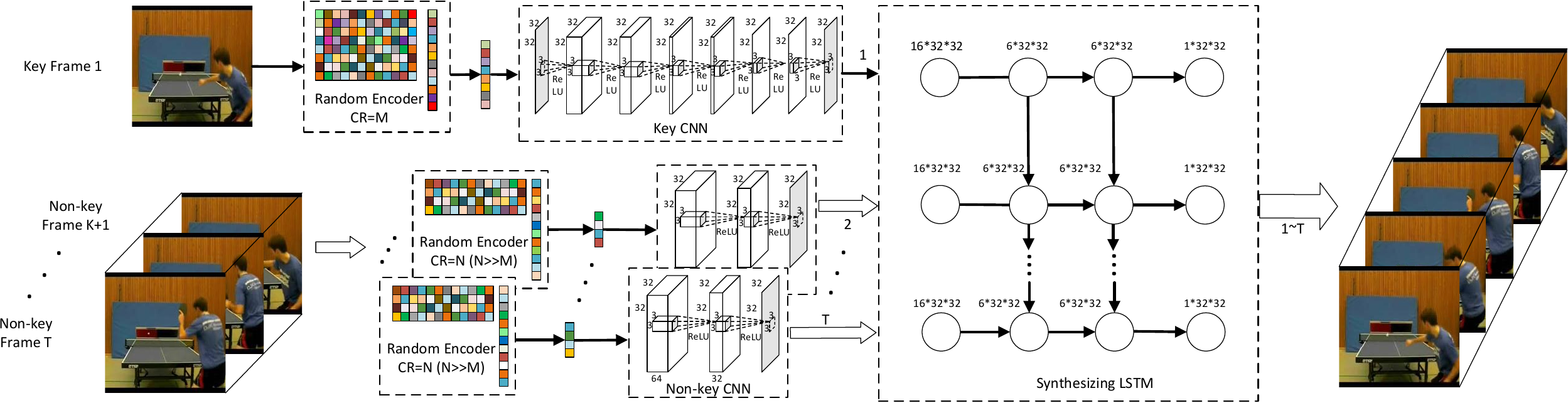}
		\caption{Overall architecture of the proposed framework. The compressed video frames are acquired by compressive sensing. In a length T GOP, the first one frame and the remaining (T-1) frames are compressed with a low and high CR, respectively. The reconstruction is performed by the CSVideoNet that is composed of a key CNN, multiple non-key CNNs, and a synthesizing LSTM.}
		\label{fig:arch}
		\vspace{-0.12in}
	\end{figure*}

	The overall architecture of the proposed video CS reconstruction framework is shown in Figure~\ref{fig:arch}. The network contains three modules: 1) an encoder (sensing matrix) for simultaneous sampling and compression; 2) a dedicated CNN for spatial features extraction after each compressed frame; 3) an LSTM for motion estimation and video reconstruction. As mentioned earlier, to improve the spatial resolution, the random encoder encodes the key frame in a GOP with more measurements and the remaining with less. Also, a recent research \cite{xu:ddcs} shows that sensing matrix can be trained with raw data to better preserve the Restricted Isometry Property (RIP). Therefore, the encoder can also be integrated into the entire model and trained with the whole network to improve reconstruction performance. Besides, as the proposed algorithm eliminates the sparsity prior constraint, the direct optimization of RIP preservation in \cite{xu:ddcs} is not necessary. Instead, we can use the reconstruction loss to train the sensing matrix along with the model. For simplicity, we still use a random Bernoulli matrix for information encoding in the experiment. Different from the prior work that extracts motion from low-resolution previews, the proposed LSTM network infers motion from high-resolution frames generated by multi-rate CNNs. The resolution of the reconstructed video is further improved with the incorporation of high-quality motion estimation.
	
	\subsubsection{Multi-rate CNN Encoder for compression ratio enhancement}
\begin{figure*} 
		\centering
		\includegraphics[width=\textwidth]{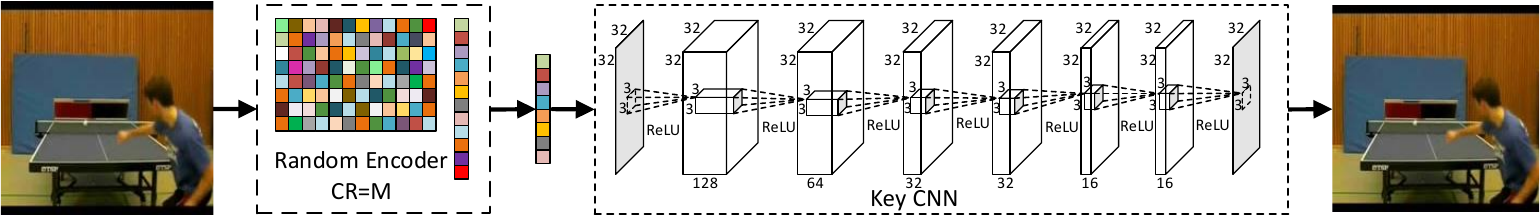}
		\caption{Pre-training of the key CNN.}
		\label{fig:TS1}
		\vspace{-0.12in}
	\end{figure*}

	Typical CNN architectures used for recognition, classification, and segmentation that map input to rich hierarchical visual features is not applicable to the reconstruction problem. The goal of the CNN is not only to extract spatial visual features but also to preserve details as much as possible. Therefore, we eliminated the pooling layer which causes information loss. Also, we discard the convolution-deconvolution architecture (widely used in segmentation tasks \cite{hye:deconv}), which first encodes salient visual features into low-dimension space and then interpolates the missing information to generate a high-resolution image. Instead, we design a special CNN suitable for CS reconstruction, which has the best recovery performance among all the tested structures mentioned above. The overall network structure is shown in Figure~\ref{fig:TS1}. All feature maps have the same dimension as the reconstructed video frames, and the number of feature maps decreases monotonically. This process resembles the sparse coding stage in CS, where a subset of dictionary atoms is combined to form the estimation of the original input. There is a fully-connected (FC) layer, denoted in gray color in Figure~\ref*{fig:TS1}, which converts vectorized $\mathit{m}$-dimensional video data to 2D features maps. To reduce the latency of the system and to simplify the network architecture, we use video blocks as input and set the block size $\mathit{n}$ to 32$\times$32. All convolutional layers are followed by a ReLU layer except the final layer. We pre-train an eight-layer $\textit{key CNN}$ to process the key frame that is compressed with a low CR. For other non-key frames compressed with a high CR, we use 3-layer $\textit{non-key CNNs}$ to handle them since they carry information of low entropy. All weights of the non-key CNNs are shared to reduce the requirement of storage. Hence the proposed framework can be easily generalized to other high-frame-rate video applications that require a larger number of non-key frames. It should be noted that the pre-training of the key CNN is critical for improving the reconstruction performance. In the case where the whole network is trained from scratch without any pre-training, the convergence performance is bad. The reason is partly due to the vanishing gradients, since we have a long path from the CNNs to the LSTM. The pre-training greatly alleviate this problem.
	
	\subsubsection{Motion-estimation synthesizing LSTM Decoder for spatial-temporal resolution enhancement}
	The proposed framework is end-to-end trainable, computationally efficient, and requires no pre/post-processing. This is achieved by performing motion estimation implicitly, which is different from prior works \cite{Aswin:multiplexing, Yang:GMM, Fowler:motion}. We utilize an LSTM network to extract motion features  that are critical for improving temporal resolution from the CNN output. Since the information flows from the first LSTM node to the remaining, the LSTM will implicitly infers representations for the hidden motion from the key frame to the non-key frames. Therefore, the recovery quality of the GOP is improved by the aggregation of motion and spatial visual features. That is why we call this network the \textit{motion-estimation synthesizing LSTM}. For simplicity, each input LSTM node in the experiment accepts input data with equal length. In fact, since the non-key frames carry less information than the key frame, the LSTM network can be designed to accept inputs with variable lengths. Hence, we can further reduce the model size and get a faster reconstruction speed. From the experiment results, we find the utilization of the LSTM network is critical to improving recovery fidelity. As a result, our model outperforms the competitive algorithms by a significant margin.
	
	The update of the LSTM units is as follows:
	\begin{eqnarray*}
		\ii_t &=& \sigma\left(W_{xi}\xx_t + W_{hi}\hh_{t-1} + W_{ci}\cc_{t-1} + \bb_i\right),\\
		\ff_t &=& \sigma\left(W_{xf}\xx_t + W_{hf}\hh_{t-1} + W_{cf}\cc_{t-1} + \bb_f\right),\\
		\cc_t &=& \ff_t \cc_{t-1} + \ii_t \tanh\left(W_{xc}\xx_t + W_{hc}\hh_{t-1} + \bb_c\right),\\
		\oo_t &=& \sigma\left(W_{xo}\xx_t + W_{ho}\hh_{t-1} + W_{co}\cc_{t} + \bb_o \right),\\
		\hh_t &=& \oo_t\tanh(\cc_t),
	\end{eqnarray*}
	where $\xx_t$ is the visual feature output of the CNN encoder. The detailed information flow and the output dimension at each LSTM node is shown in Figure~\ref{fig:arch}. The number on the LSTM nodes denotes the dimension of the output features. Specifically, the output feature map of each CNN has a dimension of 16x32x32. All these feature maps are directly fed into the input nodes of the LSTM. The LSTM has two hidden layers, the dimension of the output of each hidden layer is 6x32x32. The dimension of the final	output is 1x32x32.
	
	\subsection{Learning algorithm}   	
	Given the ground-truth video frames ${x_{\{1,\cdots,T\}}}$ and the corresponding compressed frames ${y_{\{1,\cdots,T\}}}$, we use mean square error (MSE) as the loss function, which is defined as:
	\begin{equation} 
	L(\mathbf{W,b}) = \frac{1}{2N}\sum_{i}^{T}\|f(y_i; \mathbf{W,b})-x_i\|_2^2,
	\end{equation}
	where $\mathbf{W}$, $\mathbf{b}$ are network weights and biases, respectively. 
	
	Using MSE as the loss function favors high PSNR. PSNR is a commonly used metric to quantitatively evaluate recovery quality. From the experiment results, we illustrate that PSNR is partially correlated to the perceptual quality. To derive a better perceptual similarity metric will be a future work. The proposed framework can be easily adapted to a new loss function.
	
	Three training algorithms, i.e., SGD, Adagrad \cite{Duchi:adagrad} and Adam \cite{Kingma:adam} are compared in the experiment. Although consuming most GPU memory, Adam converges towards the best reconstruction results. Therefore, Adam is chosen to optimize the proposed network. 
	
	\section{Experiment}
	As there is no standard dataset designed for video CS, we use UCF-101 dataset introduced in \cite{Soomro:ucf} to benchmark the proposed framework. This dataset consists of 13k clips and 27 hours of video recording data collected from YouTube, which belong to 101 action classes. Videos in the dataset are randomly split into 80$\%$ for training, 10$\%$ for validation and the remaining for testing. Videos in the dataset have a resolution of 320$\times$240 and are sampled at 25 fps. We retain only the luminance component of the extracted frames and crop the central 160$\times$160 patch from each frame. These patches are then segmented into 32$\times$32 non-overlapping image blocks. We get 499,760 GOPs for training and testing in total. 
	
	We set three test cases with CRs of 25, 50 and 100, respectively. Since the CR for key and non-key frames are different in the proposed method, we derive and define the CR for a particular GOP as follows. Let $m1, m2$ denotes the dimension of compressed key and non-key frame, respectively.
	Let $n$ denotes the dimension of raw frames. $T$ is the sequential length of a GOP.
	\begin{align}
		CR_1 =& n / m1, CR_2 = n / m2, \nonumber \\
		CR =& \frac{CR_1 \times 1 + CR_2  \times  (T-1)}{T}.
	\end{align}
	
	In the experiment, the CR of each key frame is m1=5, and the CR of non-key frames in each test case is m2=27, 55, and 110, respectively. Therefore, the averaged CR for each test case is about 25, 50, and 100, respectively.	
	
	The dimension of data for pre-training the key CNN is $(N \times C \times H \times W)$, where $N$=100 is the batch size, $C$=1 is the channel size, and $W, H$=(32, 32) is the height and width of each image block, respectively. The dimension of the data used for training the entire model is $(N^\prime \times T \times C \times H \times W)$, where $T$=10 is the sequence length for one GOP, and $N^\prime$=20 is the batch size. The other dimensions are the same. We shrink the batch size here because of the GPU memory limitation. In every ten consecutive video frames, we define the first one as the key frame, and the remaining as non-key frames.

	\subsection{Comparison with the state-of-the-art}
	We compare our algorithm with six reference work for CS reconstruction: \cite{Yang:GMM, Fowler:motion, Mousavi:DLRecovery, Metzler:damp, Kulkarni:ReconNet, Iliadis:Video}. We summarize all baseline approaches and our approach in Table~\ref{tab:summary}. For a fair comparison, we also re-train reference algorithms using UCF-101 dataset. Three metrics: Peak signal-to-noise ratio (PSNR), structural similarity (SSIM) \cite{Zhou:ssim}, and pixel-wise mean absolute error (MAE) are applied for performance evaluation. Note that MAE is the averaged absolute error of each pixel value within the range of [0,255], which gives a straightforward measure of the pixel-wise distortion. The authors of VCSNet only offer a pre-trained model with CR of 16, without providing sufficient training details to reproduce the experiment at present. Therefore, we train the proposed model and compare it with CVSNet at a single CR of 16.
	
	\begin{table}[]
		\centering
		\caption{Summary of major differences between the proposed approach and all baselines.}
		\label{tab:summary}
		\scalebox{0.65}{
			\begin{tabular}{c|c|c|c}
				\hline \hline
				\multirow{3}{*}{Image CS} & Iterative Based                  & \begin{tabular}[c]{@{}l@{}}Denoising-based\\ approximate message passing\end{tabular}                          & D-AMP \cite{Metzler:damp}     \\ \cline{2-4} 
				& \multirow{2}{*}{Non-iterative Based}   & Stacked denoising autoencoder                                                                                  & SDA  \cite{Mousavi:DLRecovery}      \\ \cline{3-4} 
				&                              & Convolutional neural network                                                                                   & ReconNet \cite{Kulkarni:ReconNet}  \\ \hline
				\multirow{4}{*}{Video CS} & \multirow{2}{*}{Iterative Based} & \begin{tabular}[c]{@{}c@{}}Motion-compensated block-\\ based CS with smooth\\  projected Landweber\end{tabular} & MC-BCS-SPL \cite{Fowler:motion} \\ \cline{3-4} 
				&                              & Gaussian mixture model                                                                                         & GMM \cite{Yang:GMM}       \\ \cline{2-4} 
				& \multirow{2}{*}{Non-iterative Based}   & Fully-connected neural network                                                                                 & VCSNet \cite{Iliadis:Video}    \\ \cline{3-4} 
				&                              & Proposed approach                                                                                                & \textbf{CSVideoNet} \\ \hline \hline
		\end{tabular}}
		\vspace{-0.12in}
	\end{table}
	
	\subsubsection{Comparison with image CS approaches}
	We first compare with the algorithms used for image CS reconstruction. D-AMP is a representative of the conventional iterative algorithms developed for CS, e.g., matching pursuit, orthogonal mating pursuit, iterative hard-thresholding. It offers state-of-the-art recovery performance and operates tens of times faster compared to other iterative methods \cite{Metzler:damp}. Both SDA and ReconNet are DNN-based reconstruction approaches for images proposed recently. Specifically, ReconNet is based on CNN and achieves state-of-the-art performance among all image CS reconstruction algorithms \cite{Kulkarni:ReconNet}. In the experiment, we tested both frame-based and block-based D-AMP that reconstructs an entire frame and an image block at a time, respectively. For other approaches, we test them in a block-based pattern to reduce the difficulty for training the models. The quantized results of average PSNR, SSIM, and MAE for each method under different CRs are shown in Table~\ref{tab:imageCS}. It is shown that CSVideoNet outperforms the reference approaches on all three metrics by a meaningful margin, especially at the CR of 100. The MAE of CSVideoNet is 4.59 at a 100x CR which means the averaged pixel-wise distortion is only $4.59 / 255 = 1.2\%$ compared to the ground-truth video. The PSNR drop from the CR of 25 to 100 is also calculated in Table~\ref{tab:imageCS}. We found the proposed approach suffers from the least performance degradation. This is partly due to the feature sharing between the key and non-key frames when the compressed input carries limited information.
	
\begin{figure*} []
		\centering
		\includegraphics[width=0.7\textwidth]{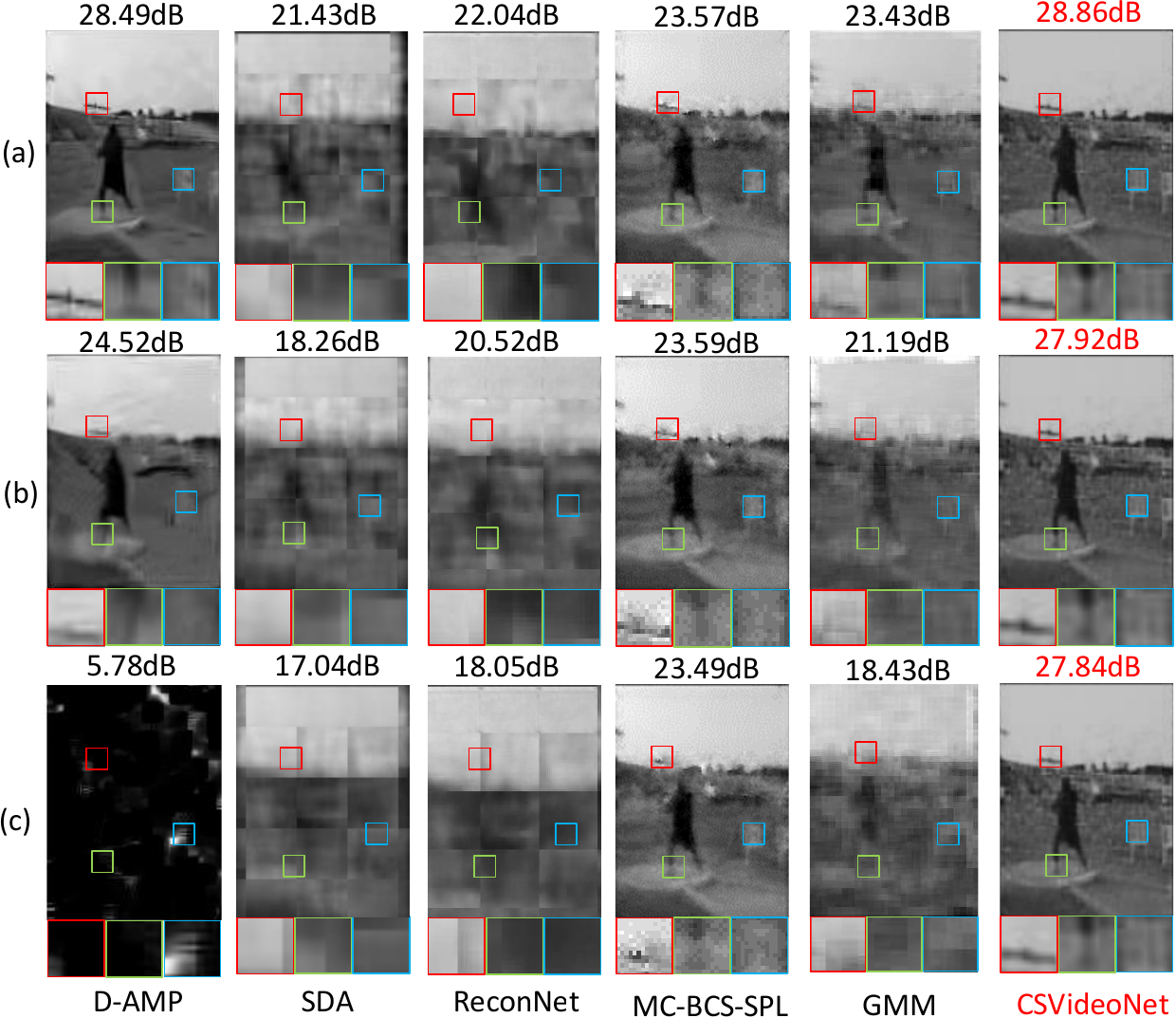}
		\caption{Illustration of reconstruction results for each method at the CR of (a) 25, (b) 50, and (c) 100, respectively.}
		\label{fig:recImage}
		\vspace{-0.12in}
	\end{figure*}

	For visual quality assessment purpose, we list the reconstructed frame by each approach in Figure~\ref{fig:recImage}. The reconstructed frame is the middle (fifth) frame in a GOP. We find all the reconstructed non-key frames have homogeneous recovery quality, and the key frame has slightly better reconstruction quality than the non-key frames. As the proportion of key and non-key frames is 1:9, and the reconstruction quality of the video is dominated by that of the non-key frames. Therefore, the middle frame (a non-key frame) shown in Figure~\ref{fig:recImage} well represents the average reconstruction quality. 
	
	For all the numerical results, we calculate all the	quality metrics, including PSNR, SSIM, and MAE, by averaging the results over all frames in a GOP. We can see that CSVideoNet provides the finest details among all approaches. The edges produced by CSVideoNet is much sharper, while such details are no longer preserved by other methods after reconstruction. This comparison demonstrates that the temporal correlation is critical for video reconstruction, the overlook of such features will significantly degrade the recovery quality of videos. Therefore, the conventional image CS approaches are not suitable for video applications.
	
	\begin{figure}[]
		\centering
		\includegraphics[width=0.35\textwidth]{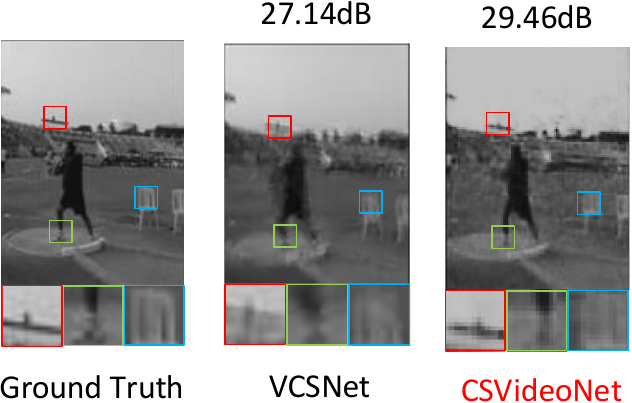}
		\caption{Illustration of reconstruction results at the CR of 16.}
		\label{fig:recVideo}
		\vspace{-0.12in}
	\end{figure}
	
	\begin{table}[]
		\centering
		\caption{Performance comparison with image CS reconstruction approaches.}
		\label{tab:imageCS}
		\scalebox{0.68}{
			\begin{tabular}{c|c|c|c|c|c|c}
				\hline \hline
				& CR  & D-AMP(F) & D-AMP(B) & SDA    & ReconNet & CSVideoNet \\ \hline
				\multirow{3}{*}{PSNR} & 25  & 25.34    & 15.1494  & 23.39  & 24.27    & \textbf{26.87}  \\ \cline{2-7} 
				& 50  & 12.49    & 9.1719   & 21.96  & 22.47    & \textbf{25.09}  \\ \cline{2-7} 
				& 100 & 7.17     & 8.0942   & 20.40  & 20.44    & \textbf{24.23}  \\ \hline
				\multirow{3}{*}{SSIM} & 25  & 0.76     & 0.0934   & 0.69   & 0.73     & \textbf{0.81}   \\ \cline{2-7} 
				& 50  & 0.08     & 0.0249   & 0.65   & 0.67     & \textbf{0.77}   \\ \cline{2-7} 
				& 100 & 0.03     & 0.0067   & 0.61   & 0.61     & \textbf{0.74}   \\ \hline
				\multirow{3}{*}{MAE}  & 25  & 4.65     & 24.92    & 5.76   & 5.02     & \textbf{3.38}   \\ \cline{2-7} 
				& 50  & 64.30    & 81.67    & 6.60   & 5.67     & \textbf{4.31}   \\ \cline{2-7} 
				& 100 & 92.12    & 86.04    & 8.50   & 7.42     & \textbf{4.59}   \\ \hline
				PSNR$\downarrow$           & 25 $\rightarrow$ 100 & 72$\%$  & 13$\%$  & 47$\%$     & 16$\%$     & \textbf{10$\%$}       \\ \hline \hline
		\end{tabular}}
	\vspace{-0.12in}
	\end{table}
	
	\subsection{Comparison with video CS approaches}
	We compare the proposed CSVideoNet with existing video CS approaches. MC-BCS-SPL estimates motion directly from the current and the reference frame. GMM models the spatial-temporal correlation by assuming all pixels within a video patch are drawn from a GMM distribution. GMM has the state-of-the-art performance among conventional model-based video CS approaches \cite{Yang:GMM}. To the best of our knowledge, \cite{Iliadis:Video} is the only DNN-based work proposed for video CS. The quantized results of average PSNR, SSIM, and MAE for each method under different CRs are shown in Table~\ref{tab:videoCS}. It is observed that the proposed approach improves PSNR by 3 to 5dB over the reference methods. Specifically, we find MC-BCS-SPL and GMM have similar performance and perform much better than the model-based image CS approach, D-AMP. However, their performance are similar to SDA and ReconNet, which are designed for processing images. This implies that the conventional model-based methods suffer from limited performance due to the limited model capacity when dealing with large-scale problem. Even though they consider the temporal correlation among video frames, the model capacity is insufficient for visual patterns. To improve performance, one could increase the size of the conventional models. However, the computational complexity forof these meods will also increase substantially, inhibiting their application to video CS.
	
	\begin{table}[]
		\centering
		\caption{Performance comparison with video CS reconstruction approaches.}
		\label{tab:videoCS}
		\scalebox{0.8}{
			\begin{tabular}{c|c|c|c|c}
				\hline \hline
				& CR  & MC-BCS-SPL & GMM   & CSVideoNet \\ \hline
				\multirow{3}{*}{PSNR}   & 25  & 22.41      & 23.76 & \textbf{26.87}      \\ \cline{2-5} 
				& 50  & 20.59      & 21.26 & \textbf{25.09}      \\ \cline{2-5} 
				& 100 & 19.67      & 19.64 & \textbf{24.23}      \\ \hline
				\multirow{3}{*}{SSIM}   & 25  & 0.37       & 0.72  & \textbf{0.81}       \\ \cline{2-5} 
				& 50  & 0.30       & 0.61  & \textbf{0.77}       \\ \cline{2-5} 
				& 100 & 0.19       & 0.54  & \textbf{0.74}       \\ \hline
				\multirow{3}{*}{MAE} & 25  & 11.88      & 5.14  & \textbf{3.38}       \\ \cline{2-5} 
				& 50  & 16.03      & 7.50  & \textbf{4.31}       \\ \cline{2-5} 
				& 100 & 28.86      & 9.37  & \textbf{4.59}       \\ \hline 
				PSNR$\downarrow$   & 25 $\rightarrow$ 100 & 26$\%$  & 17$\%$ & \textbf{10$\%$} \\ \hline \hline
		\end{tabular}}
	\vspace{-0.12in}
	\end{table}
	
	DNN provides a viable solution. Both CSVideoNet and VCSNet are designed for video CS reconstruction. For reasons explained earlier, we compare the two approaches at a CR of 16. The results are shown in Table~\ref{tab:VNet} and Figure~\ref{fig:recVideo}. Both the two approaches achieve high recovery quality compared to other baselines. However, VCSNet is a plain fully-connect network that has limited capability for processing sequential data. As a result, it suffers from a low-quality motion estimation, which explains why it has inferior performance compared to the proposed solution.
	
	\begin{table}[]
		\centering
		\caption{Performance comparison with VCSNet at the CR of 16.}
		\label{tab:VNet}
		\scalebox{0.8}{
			\begin{tabular}{c|c|c}
				\hline \hline
				& VCSNet   & CSVideoNet 	  \\ \hline
				PSNR  & 25.07704 & 28.078     \\ \hline
				SSIM  & 0.817669 & 0.8431     \\ \hline
				MAE & 3.887867 & 2.9452       \\ \hline \hline
		\end{tabular}}
	\vspace{-0.12in}
	\end{table}
	\begin{table}[]
		\centering
		\caption{Structures of CNN1 and CNN2.}
		\label{tab:strCNN}
		\begin{threeparttable}
			\scalebox{0.68}{
				\begin{tabular}{c|c|c|c|c|c|c|c|c|c|c|c|c|c}
					\hline \hline
					\# Layer & 1 & 2   & 3   & 4   & 5   & 6   & 7  & 8  & 9  & 10 & 11 & 12 & 13 \\ \hline
					CNN1     & 1 & 128 & 64  & 32  & 32  & 16  & 16 & 1  &    &    &    &    &    \\ \hline
					CNN2     & 1 & 512 & 256 & 256 & 128 & 128 & 64 & 64 & 32 & 32 & 16 & 16 & 1  \\ \hline \hline
				\end{tabular}
			}
			\begin{tablenotes}\footnotesize
				\tiny \item[*] CNN1 is used in CSVideoNet. The dimension of all feature maps in both CNNs are 32$\times$32.
			\end{tablenotes}
		\end{threeparttable}
	\end{table}
	
	To illustrate that the performance improvement of the proposed approach comes from integrating temporal features through the LSTM network rather than simply increasing the model size, we set another experiment, in which we compare the performance of two CNNs with different sizes. The structure of the two CNNs are shown in Table~\ref{tab:strCNN}, and the performance comparison is shown in Table~\ref{tab:cpCNN}. We can see that simply increasing the size of CNN does not provide meaningful improvement for reconstruction. This, wh be explained by the incapability of CNN to capture temporal features. The incorporation of the LSTM network improves the PSNR by up to 4 dB, which represents more than twice of error reduction. Specifically, the performance improvement increases with thealong wiachieves theits maximum wheR is 100. This explains that the implicit motion estimation by LSTM is critical to the video CS reconstruction especially at high CRs. 

	\begin{table}[]
		\centering
		\caption{Runtime comparison for reconstructing a 160$\times$160 video frame at different CRs.}
		\label{tab:tab4}
		\resizebox{0.30\textwidth}{!}{%
			\begin{tabular}{c|c|c|c}
				\hline \hline
				Model      & CR=25  & CR=50  & CR=100 \\ \hline
				D-AMP(F)   & 38.37  & 41.20  & 31.74  \\ \hline
				D-AMP(B)   & 8.4652 & 8.5498 & 8.4433 \\ \hline
				SDA        & 0.0278 & 0.027  & 0.023  \\ \hline
				ReconNet   & 0.064  & 0.063  & 0.061  \\ \hline
				MC-BCS     & 7.17   & 8.03   & 9.00   \\ \hline
				GMM        & 8.87   & 10.54  & 18.34  \\ \hline
				CSVideoNet & 0.0094 & 0.0085 & 0.0080 \\ \hline \hline
			\end{tabular}
		}
	\vspace{-0.1in}
	\end{table}

	\begin{table}[]
		\centering
		\caption{Performance comparison with CNN methods.}
		\label{tab:cpCNN}
		\scalebox{0.75}{
			\begin{tabular}{c|c|c|c|c}
				\hline \hline
				& CR  & CNN1  & CNN2  & CSVideoNet \\ \hline 
				\multirow{3}{*}{PSNR}  & 25  & 24.27 & 23.74 & 26.87      \\ \cline{2-5} 
				& 50  & 22.47 & 22.17 & 25.09      \\ \cline{2-5} 
				& 100 & 20.44 & 20.10 & 24.23      \\ \hline
				\multirow{3}{*}{SSIM}  & 25  & 0.73  & 0.69  & 0.81       \\ \cline{2-5} 
				& 50  & 0.67  & 0.65  & 0.77       \\ \cline{2-5} 
				& 100 & 0.61  & 0.58  & 0.74       \\ \hline
				\multirow{3}{*}{MAE} & 25  & 5.02  & 6.46  & 3.38       \\ \cline{2-5} 
				& 50  & 5.67  & 6.23  & 4.31       \\ \cline{2-5} 
				& 100 & 7.42  & 8.92  & 4.59       \\ \hline \hline
			\end{tabular}
		}
	\vspace{-0.18in}
	\end{table}

	\subsection{Performance under noise}
\begin{figure}[]
		\centering
		\includegraphics[width=0.27\textwidth]{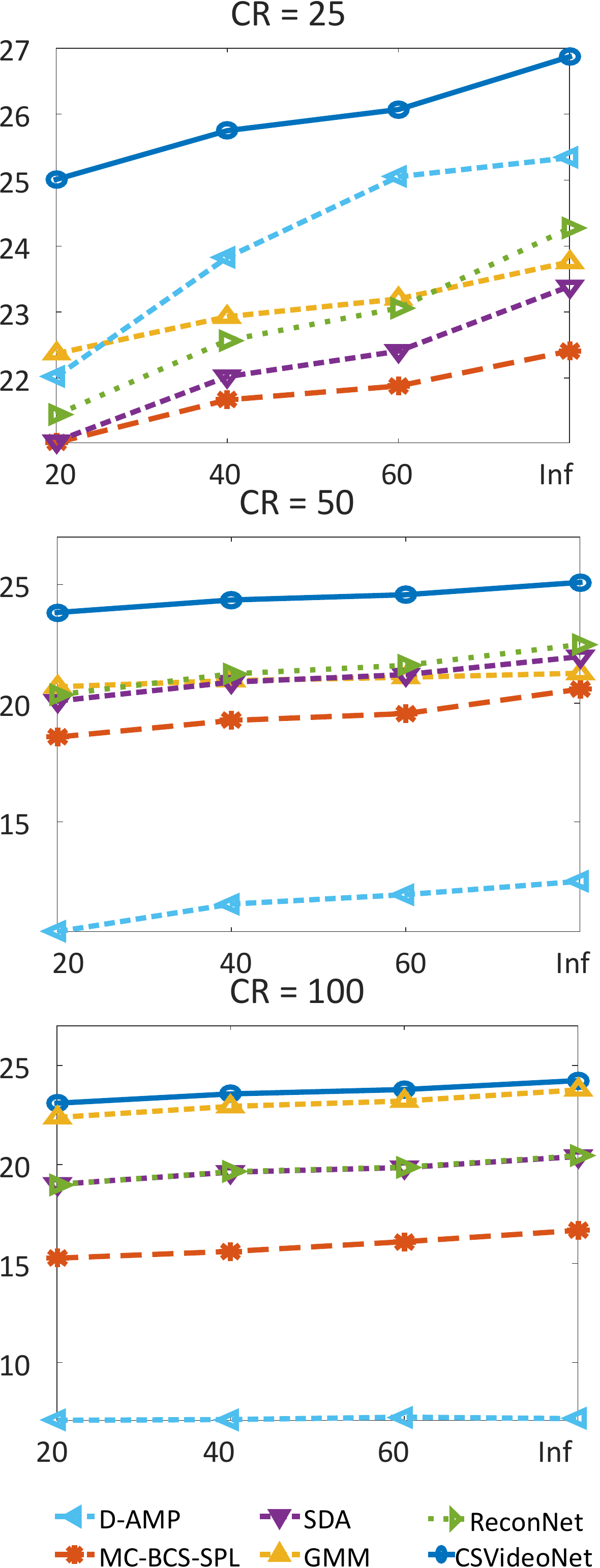}
		\caption{PSNR comparison at different SNRs.}
		\label{fig:noise}
		\vspace{-0.20in}
	\end{figure}

	To demonstrate that the robustness of CSVideoNet to sensor noise, we conduct a reconstruction experiment with input videos contaminated by random Gaussian noise. In this experiment, the architecture of all DNN-based frameworks remains the same as in the noiseless case. We test the performance at three levels of SNR - 20dB, 40dB, and 60dB. For each noise level, we evaluate all approaches at three CRs of 25, 50, and 100. The average PSNR\ achieved by each method at different CRs and noise levels are shown in Figure~\ref{fig:noise}. It can be observed that CSVideoNet can reliably achieve a high PSNR across at different noise levels and outperform the reference methods consistently.

	\subsection{Time complexity}
	We benchmark the runtime performance of different methods. Due to the iterative nature of conventional CS algorithms (D-AMP, MC-BCS-SPL, GMM), they suffer from high data-dependency and low parallelism, which is not suitable for GPU acceleration. Due to the lack of GPU solvers, we run these reference algorithms on an octa-core Intel Xeon E5-2600 CPU. Benefiting from the feedforward data-path and high data concurrency of DNN-based approaches, we accelerate CSVideoNet and other DNN-based baselines using a Nvidia GTX Titan X GPU. The time cost for fully reconstructing a video frame in the size of (160$\times$160) are compared in Table~\ref{tab:tab4}. CSVideoNet consumes 8 milliseconds (125 fps) to reconstruct a frame at the CR of 100. This is three orders of magnitude faster than the reference methods based on iterative approaches. The time cost of VCSNet and CSVideoNet at the CR of 16 is 3.5 and 9.7 milliseconds, respectively. Through further hardware optimization, we believe CSVideoNet has the potential to be integrated into CS cameras to enable the real-time reconstruction of high-frame-rate video CS.
	\section{Conclusion}
	\vspace{-0.03in}
	In this paper, we present a real-time, end-to-end, and non-iterative framework for high-frame-rate video CS. A multi-rate CNN variant and a synthesizing LSTM network are developed to jointly extract spatial-temporal features. This is the key to enhancing the compression ratio and recovery quality trade-off. The magnificent model capacity of the proposed deep neural network allows to map the inverse mapping of CS without exploiting any sparsity constraint. The feed-forward and high-data-concurrency natures of the proposed framework are the key to enabling GPU acceleration for real-time reconstruction. Through performance comparison, we demonstrate that CSVideoNet has the potential to be extended as a general encoding-decoding framework for high-frame-rate video CS applications. In the future work, we will exploit the effective learning methods to decode high-level information from compressed videos, e.g., object detection, action recognization, and scene segmentation.
	
	\section{Acknowledgement}
	This work is supported by NSF grant IIS/CPS-1652038. The research infrastructure used by this work is supported by NSF grant CNS-1629888.
	{\small	
		\bibliographystyle{abbrv}
		\bibliography{references}
	}
	
\end{document}